\documentclass[10pt,twocolumn,letterpaper]{article}

\usepackage{iccv}
\usepackage{times}
\usepackage{epsfig}
\usepackage{graphicx}
\usepackage{amsmath}
\usepackage{amssymb}

\usepackage{iccv}
\usepackage[dvipsnames]{xcolor}

\usepackage{multirow}

\usepackage[breaklinks=true,bookmarks=false]{hyperref}

\iccvfinalcopy 

\def\iccvPaperID{1692223361278} 

\ificcvfinal\fi

\begin{document}

\title{MARL: Multi-scale Archetype Representation Learning for Urban Building Energy Modeling}

\author{Xinwei Zhuang*, Zixun Huang*, Wentao Zeng, Luisa Caldas\\
University of California, Berkeley\\
{\tt\small \{xinwei\_zhuang, zixun, wentao\_zeng, lcaldas\}@berkeley.edu}
}

\maketitle    
\def\thefootnote{*}\footnotetext{Equal Contribution}
\def\thefootnote{\arabic{footnote}}
\ificcvfinal\thispagestyle{empty}\fi

\begin{abstract}
Building archetypes, representative models of building stock, are crucial for precise energy simulations in Urban Building Energy Modeling. The current widely adopted building archetypes are developed on a nationwide scale, potentially neglecting the impact of local buildings’ geometric specificities. We present Multi-scale Archetype Representation Learning (MARL), an approach that leverages representation learning to extract geometric features from a specific building stock. Built upon VQ-AE, MARL encodes building footprints and purifies geometric information into latent vectors constrained by multiple architectural downstream tasks. These tailored representations are proven valuable for further clustering and building energy modeling. The advantages of our algorithm are its adaptability with respect to the different building footprint sizes, the ability for automatic generation across multi-scale regions, and the preservation of geometric features across neighborhoods and local ecologies. In our study spanning five regions in LA County, we show MARL surpasses both conventional and VQ-AE extracted archetypes in performance. Results demonstrate that geometric feature embeddings significantly improve the accuracy and reliability of energy consumption estimates. Code, dataset and trained models are available on the project page: \href{https://github.com/ZixunHuang1997/MARL-BuildingEnergyEstimation}{https://github.com/ZixunHuang1997/MARL-BuildingEnergyEstimation}. 
\end{abstract}

\section{Introduction}
The built environment plays a significant role in global climate change, accounting for approximately 34\% of worldwide final energy consumption in 2020 \cite{unep2022globalstatus}. In light of the urgent worldwide issues concerning sustainability and energy efficiency, there has been a concentrated focus on research in Urban Building Energy Modeling (UBEM) \cite{ali2021review}. UBEM has evolved into a crucial computational tool that enables the analysis and forecast of energy consumption at an urban scale. Given the impracticability or infeasibility of modeling each individual building in minute detail, developing building archetypes for the study area has become a critical determinant for precision for urban scale energy estimation. These archetypes represent specific groups of buildings sharing similar characteristics and energy performance within the area of interest. Their strategic use in UBEM allows for efficient and effective energy modeling at the urban scale.

Recent studies have established global building archetypes \cite{doe_prototype, loga2016tabula, reyna2022us}. However, many of these are designed at a country scale, depend on expert input \cite{doe_prototype}, and don't integrate actual building geometry, leading to potential inaccuracies and limiting their scalability, particularly in areas where computational resources and data accessibility present challenges. Additionally, urban-scale energy estimation demands significant computational resources, often inaccessible to disadvantaged communities and neighborhoods. Such disparities can introduce data biases, curtail global applicability, and limit the UBEM's diversity mainly to major cities like Boston \cite{cerezo_davila_modeling_boston_2016, chen_citybes_development_2019} and San Francisco \cite{chen_citybes_development_2019}. These limitations hinder the effectiveness of building archetypes in energy-efficient design and implementation.

\begin{figure}[t]
\begin{center}
   \includegraphics[width=\linewidth]{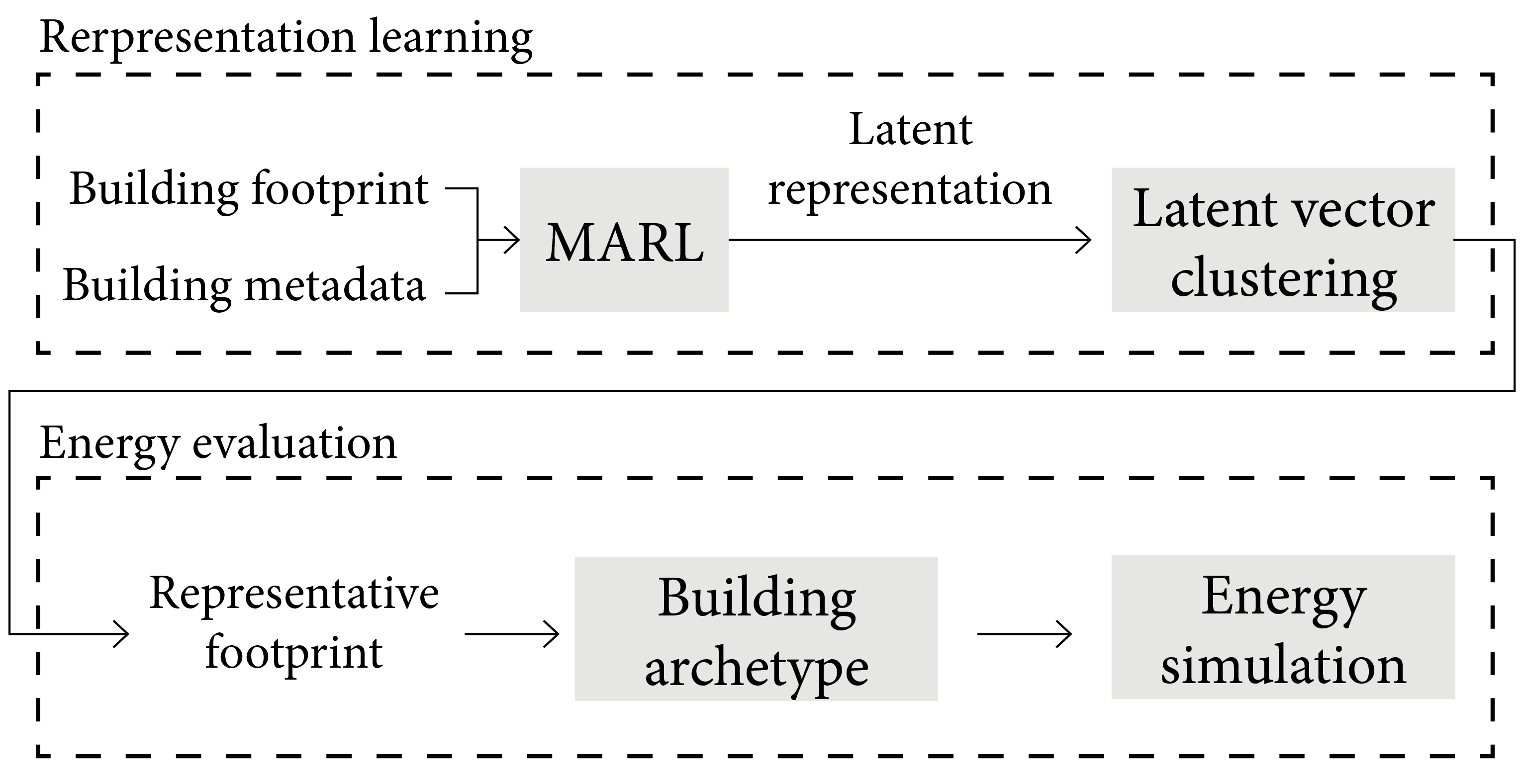}
\end{center}
   \caption{\textbf{Building Archetype Finding Pipeline.} Besides data collection, our method consists of 2 stages: archetype representation learning and clustering, and energy simulation.}
\label{fig:long}
\label{fig:workflow_overall}
\end{figure}

In response, we present Multi-scale Archetype Representation Learning (MARL,  Figure \ref{fig:workflow_overall}), a method designed to automate local building archetype construction through representation learning. Our proposed method addresses the aforementioned challenges by refining the essential elements of building archetypes for UBEM. 
Our main contributions are summarized as follows:
\begin{itemize}
\item We introduce building geometry into the building archetype construction process.
\item We provided a cost-effective solution for Urban Building Energy Modeling (UBEM).
\item We propose an image reconstruction-based framework that streamlines the archetype creation process. 
\item We introduce the integration of downstream tasks into the building archetype construction process, which increases the accuracy of our models, thereby enabling more precise and reliable energy consumption estimates at an urban scale.
\item We conduct open-set experiments to demonstrate that our trained model possesses transferable and reusable characteristics in similar urban and geographical environments.
\end{itemize}

\section{Related Studies}
\subsection{Representation learning}
Representation learning offers effective strategies for dealing with high-dimensional data forms such as images and 3D models. Integral to these methods are a variety of neural compression algorithms \cite{yang_introduction_2022}, such as normalizing flows \cite{rezende_variational_2016}, variational autoencoders \cite{kingma_auto-encoding_2013}, diffusion probabilistic models \cite{ho_denoising_2020}, and generative adversarial networks \cite{goodfellow_generative_2014}. Recent advancements have also seen variations of these models, such as Vector Quantized Variational Autoencoders (VQ-VAEs) \cite{oord2018neural} and Auto Decoders \cite{park2019deepsdf}. These techniques leverage the latent space as a compact representation where the high-dimensional data is encoded into lower-dimensional vectors, preserving the underlying features. A wealth of recent research showcases various application areas ranging from natural language processing to image classification \cite{devlin_bert_2019, hinton_deep_2012, krizhevsky_imagenet_2012}.
\subsection{Building archetype for energy estimation}

Building archetypes are a representative subset of buildings to model an entire building stock, and are critical to the process of Urban Building Energy Modeling (UBEM). The prevailing methodologies for constructing the representative building archetype leverage statistical analysis or use generalized assumptions informed by expert knowledge \cite{cerezo2016modeling, deru2011us}, and remain a time-consuming task \cite{deru2011us}. Existing archetypes define representative buildings based on categorical attributes such as construction material, vintage, area, stories, and energy system \cite{doe_prototype, reyna2022us}. Nevertheless, these models usually do not incorporate the actual geometries of buildings into the construction of building archetypes, thus potentially overlooking the nuanced variations and distributions of prevailing geometries among different building stocks. 


Recent years have witnessed exploration into data-driven methods for the creation of building archetypes \cite{ali2021review, nutkiewicz2021exploring, reyna2016defining}, but most of these methodologies continue to exclude building geometry parameters, largely as a result of technical intricacies \cite{cerezo2016modeling}. Instead, the building geometry is simplified to numerical data, such as areas and shape ratio, with limited studies directly addressing the actual geometry \cite{dejaeger2020building}. Despite the significant impact of building geometry on energy consumption (\cite{cerezo2016modeling, coffey2015epidemiological}, its exclusion raises the potential for inaccuracies and unreliability for energy estimation in UBEM. This underscores the necessity for methods capable of incorporating building geometry to represent building stock comprehensively, thereby augmenting the precision and granularity of locally tailored building archetypes.

\section{Method}
\begin{figure*}[t]
\begin{center}
   \includegraphics[width=\linewidth]{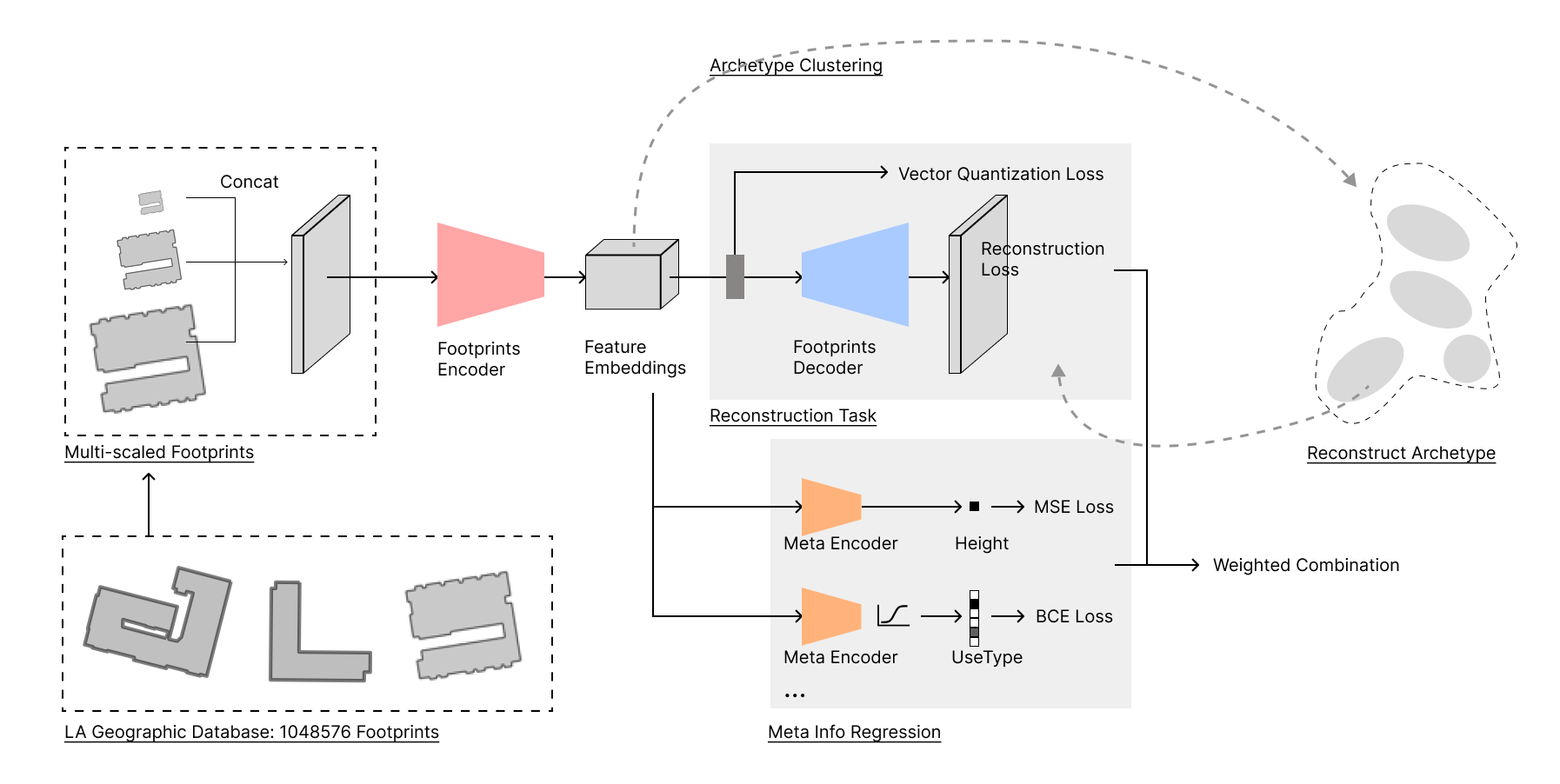}
\end{center}
   \caption{\textbf{Model Architecture for multi-scale archetype learning and clustering.} This stage consists of one auto-encoder and multiple downstream tasks.}
\label{fig:long}
\label{fig:workflow}
\end{figure*}

Besides data collection, our method consists of 2 stages: archetype representation learning and clustering, and energy simulation. Our method follows the bottom-up approach for constructing a building archetype model \cite{reyna2016defining} with a representation learning module to characterize the building stock. Multi-scale Architectural Representation Learning (MARL) is developed to encapsulate the implicit features of the input building stock into a compact, reduced-dimensional space, i.e., the latent space. Following this, we apply k-means clustering in the latent space to identify the representative building footprints, which are then converted into to building archetypes with the same parameters documented in PBM \cite{doe_prototype}. 

We then conduct energy simulations on PBM and our building archetypes and compare the aggregated energy profiles. The aim is to investigate whether utilizing real-world building geometries for a specific district enhances the accuracy of urban-scale energy estimation and to evaluate the extent to which the accuracy of such energy modeling can be improved. 

\subsection{Archetype learning and clustering}
Our model for representation learning (Figure. \ref{fig:workflow}) consists of an auto-encoder and an optional downstream task pool (DTP). The auto-encoder aims at compressing and purifying the geometric contours of the building, while the downstream tasks add a restriction and punishment on it to conserve geometric information related to vital building properties. After the representation learning is finished, we then feed all footprints in a certain region back into our trained encoder and get the clustering centers of the corresponding latent vectors to acquire archetypes in this region.

\subsubsection{Footprint reconstruction}
Since the sizes of different footprints vary widely, to ensure that our model is flexible enough in its choice of data about building contours at different scales, a single footprint is scaled into 3 different sizes. Detailed implementation is described in Section \ref{image_scale}. Both the encoder and decoder for the image reconstruction task are built with a CNN-based structure with residual connections. The scaled footprints are fused together and encoded into a latent vector with a dimension of $28\times28\times32$ by the encoder. A vector quantizer \cite{oord2018neural} is then used to decouple the implicit representation and map it to discrete embedding vectors (a code book) prior to forwarding the vector to the decoder. During the quantization step, each continuous vector output from the encoder is matched to the nearest vector in the code book, which is then forwarded to the decoder. This helps to better control the quality and diversity of the generated data based on the clustering centers in further steps.

\begin{figure}[h]
\begin{center}
   \includegraphics[width=\linewidth]{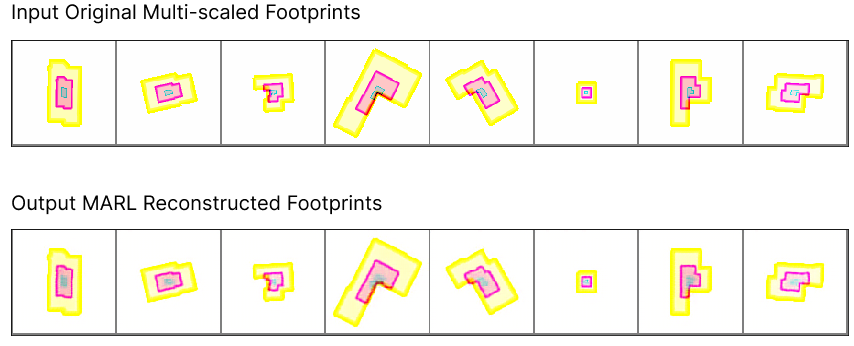}
\end{center}
   \caption{Examples of MARL footprint reconstruction.}
\label{fig:recon}
\end{figure}

Based on VQ-AE, our image reconstruction process incorporates both reconstruction loss and vector quantization loss in the proposed representation learning. 
The training objective for the reconstruction part becomes equation \ref{eq:recon}:
\begin{equation}
L = log (p(x|z_{q}(x))) + \left \| sg(z_{e}(x))-e \right \|_2^2
\label{eq:recon}
\end{equation}
The first term corresponds to the reconstruction loss, while the second term represents the codebook loss. Here, we use the notations $x$, $z_q(x)$, $z_e(x)$, and $e$ to refer to the inputs, decoder inputs, encoder outputs, and embedding vectors, respectively. And the operator $sg$ stands for the stop gradient operator.

\subsubsection{Downstream task pool}

Supervised by a downstream task pool (DTP), our MARL model excels in producing a more valuable latent space compared to normal auto-encoders, making it ideal for architectural purpose-driven applications. By leveraging the insights from the additional tasks, our model is capable of preserving implicit geometry features closely related to crucial meta information of buildings. 

The primary task focuses on predicting the building's program or purpose, while the secondary task involves predicting the 'vintage' of a building, defined as the year it was constructed. Some other simple tasks are incorporated as well to gain further constraints on the encoder, such as predicting building heights from image grayscales. All tasks are meticulously tailored and hold a strong connection to traditional building energy modeling. Moreover, they directly or indirectly influence and even shape architectural geometry to a significant extent. 

The first task aims at predicting the building program, specifically within the context of residential categories in Los Angeles. These categories are manifold, spanning from Mobile Homes, Units, and Rooming Houses to Apartments. The wide scope of categories caters to the diverse types of residential architecture prevalent within the region. For the second task, we operationalize the concept of 'vintage' as a sequence of categorical variables that encapsulate distinct temporal spans: pre-1980, 1980-2004, 2004-2013, and 2013 to the present. These divisions correspond with significant revisions to the California Building Code, serving as crucial epochs for analyzing the energy performance of structures. This segmentation is used for energy performance analysis \cite{IMMMdata}. 

By incorporating these tasks, our model is expected to gain a more nuanced understanding of building properties, thereby contributing to more precise and sophisticated predictions. These tasks reinforce the model's robustness and increase its applicability across a range of building types.

\subsection{Energy Simulation}
After generating the representative footprints, we extrude them to their corresponding heights, which are coded in greyscale. We then apply energy parameters, including the window-wall ratio, U-value of the envelope, heating type, etc., consistent with the PBM \cite{doe_prototype}. Subsequently, we employ Climate Studio, an EnergyPlus plug-in for Rhino3D, to simulate the annual energy use intensity (EUI) of each representative building under the LA airport climate data sourced from \cite{climateonebuildingorg}. After this, we aggregate the annual EUIs for each cluster based on their respective total areas and sum these to calculate the cumulative EUI for the entire neighborhood. 


\section{Experiments}

\subsection{Geographic dataset}

\begin{figure}[t]
\begin{center}
   \includegraphics[width=0.9\linewidth]{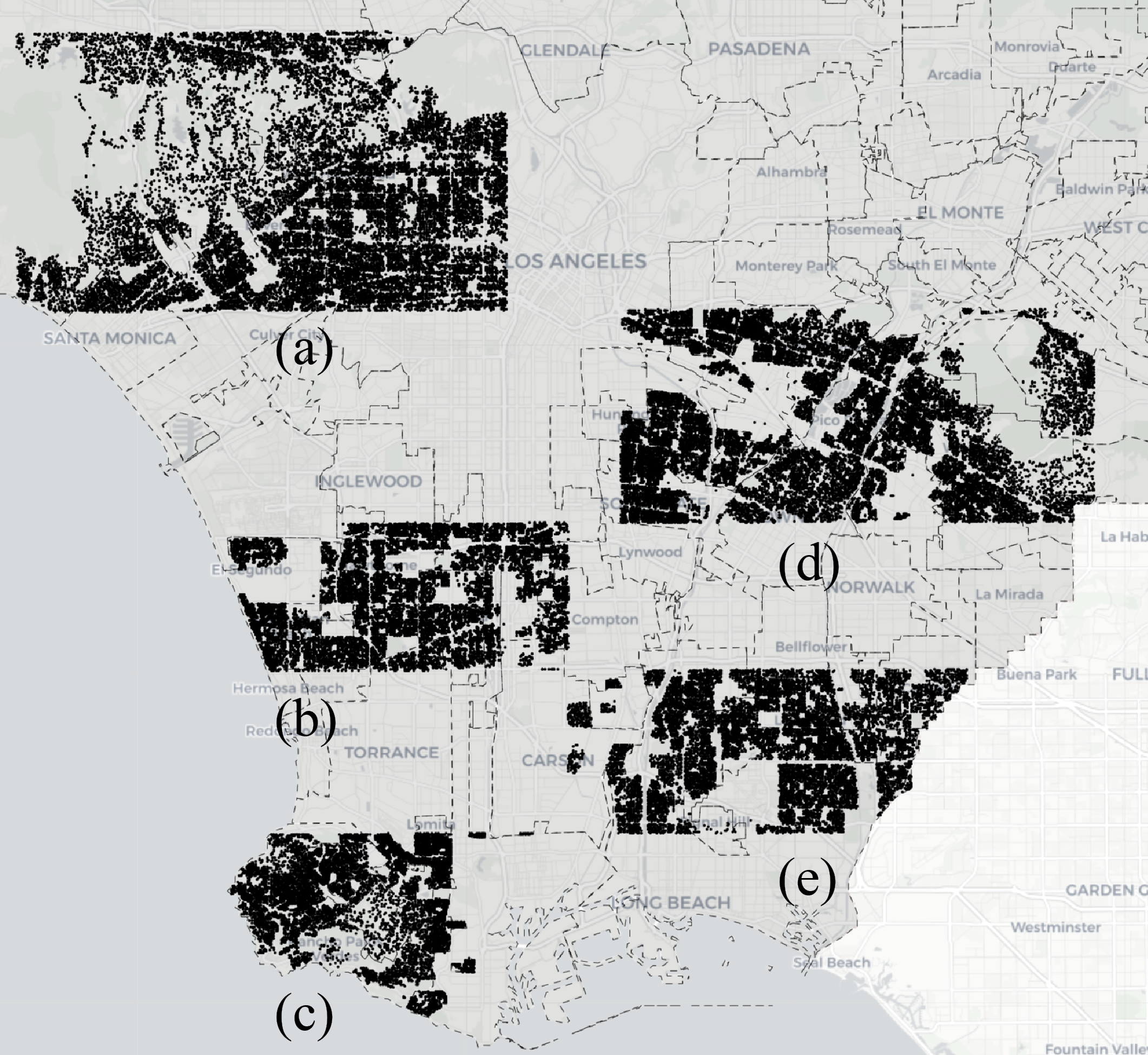}
\end{center}
   \caption{\textbf{Dataset Visualization.} Selected Neighborhood in Los Angeles County. (a) Santa Monica, Westwood, Brent Wood, Beverly Hills, west Hollywood (b) Manhattan Beach, Hawthorne, El Segundo, Lawndale Redondo Beach and Gardena (c) Rancho Palos and Rolling Hills (d) Downey, south gate, Commerce, East Los Angeles, Pico Rivers, Whitter
(e) Long Beach and Lakewood}
\label{fig:region_map}
\end{figure}

\begin{figure}[t]
\begin{center}
   \includegraphics[width=0.9\linewidth]{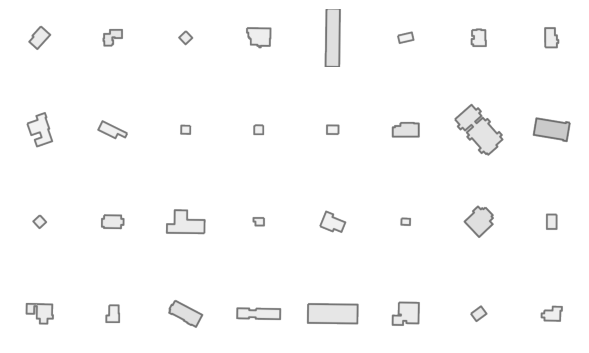}
\end{center}
   \caption{Examples of the building footprints.}
\label{fig:footprints}
\end{figure}

We selected five distinct neighborhoods within Los Angeles County, each encompassing an area in excess of 100 km\textsuperscript{2} and comprising a dense residential building population surpassing 10,000 units. These specific areas have been graphically illustrated in Figure \ref{fig:region_map}. The building footprint, alongside other building metadata including vintage, height, and building type, are derived from the Assessor Parcels Data provided by Los Angeles County \cite{countybuilding2014}. In the interest of maintaining focus on residential aspects, we narrowed down our dataset to exclusively incorporate buildings designated for residential usage (single and multi-family buildings). Building heights were normalized into the range of 0-255 and translated into greyscale values. This color-coded scheme allows for intuitive visualization of height distribution within residential buildings. Representative samples of these building footprints are showcased in Figure \ref{fig:footprints}.


Region (b), (c), and (e) each consist of 6959, 7736, 6064 footprints. Among them, 90\% of randomly selected footprints are utilized for the training process of our model, while the remaining portion is allocated for validation. Regions (a) and (d) are held out and encompass 9870, 11767 footprints each, intended for conducting open-set experiments.

\subsection{Implementation details and Metrics}\label{image_scale}
A single footprint with dimensions of $1410\times1410$ is initially divided into three sizes: $700\times700$, $224\times224$, and $112\times112$. Subsequently, all of these sizes are uniformly resized to $112\times112$ before being combined into a concatenated format.

The image embedding network consists of a 3-layer CNN and 1 residual block serving as the encoder, along with a vector quantizer, 3 up-sampling layers, and 1 residual block as the decoder. All meta-information encoders in the downstream task pool are constructed with only 1 convolution layer and 1 fully-connected layer. Feature embedding fed into both the vector quantizer and the downstream task pool is of dimension $28\times28\times32$.

Regarding the training strategy for our MARL integrated with DTP, we adopted a pre-training phase focusing solely on the reconstruction loss. Following this, we fully fine-tune the entire model's weights for one epoch through a weighted sum of the reconstruction loss and the downstream task losses.

\textbf{Ground truth. }
Given the privacy concerns, obtaining real-world data regarding building energy consumption poses a challenge. As a workaround, we used a simulated dataset, the Integrated Multisector Multiscale Modeling (IMMM) data \cite{IMMMdata}, as our ground truth for energy estimation. This dataset presents a thorough simulation of energy consumption for every building across the entire Los Angeles County.

\textbf{Baseline. }
For comparison, we adopt two prototype models from Prototype Building Models (PBM) set forth by \cite{doe_prototype}: single-family housing model and multi-family housing model under climate zone 3C for Los Angeles County. The PBM is a well-established building archetype dataset applicable nationwide and adaptable to various climate conditions. We conduct energy simulations for both single and multi-family models. The energy consumption estimation for the baseline is the sum of the product of energy consumption and the respective area for both single-family and multi-family housing models.

\textbf{Metrics. }
To evaluate the efficacy of the proposed algorithm, we use the accuracy of the aggregated energy consumption for a selected area as the evaluation metrics \eqref{equ: metric}. The absolute error $AE$ is denoted as the difference between the total estimated energy consumption $EC_{est}$ and the ground truth $EC_{gt}$.
With the building stock clustered and the representative footprint selected, create a building archetype model based on these footprints, supplementing with energy-related parameters that are identical to PBM. In this way, the only variable parameter remains the building geometry. Following this, we conduct energy simulations for each archetype model and aggregate the energy consumption results by the total area the cluster represents. The resulting aggregated energy consumption is compared with the ground truth and the energy baseline in Table \ref{sheet:multiple}.
\begin{gather}
    AE = \left |  EC_{est} - EC_{gt}\right |\\
    Accuracy = 1 - \frac{AE}{EC_{gt}}
\label{equ: metric}
\end{gather}

\begin{table*}[h!]
\begin{center}
\begin{tabular}{|l|c|c|c|c|c|}
\hline
\multicolumn{2}{|l|}{Archetype offered by} & EUI $(kWh/m^2)$ & Building Area $(m^2)$ & Energy $(kWh)$ & Accuracy $(\%)$ \\
\hline\hline
\multirow{2}*{PBM \cite{doe_prototype}} & MFH & 75.14 & 861123.64 & \multirow{2}*{137344567} & \multirow{2}*{71.62} \\
\cline{2-4}
~ & SFH & 60.79 & 1194889.74 &  &  \\
\hline

\multirow{2}*{MARL (Ours)} & MFH & 89.3 & 861123.64 & \multirow{2}*{179539369} & \multirow{2}*{93.62} \\
\cline{2-4}
~ & SFH & 85.9 & 1194889.74 &  &  \\
\hline

MARL + DTP & MFH & 92.5 & 861123.64 & \multirow{2}*{183609344} & \multirow{2}*{\textbf{95.74}} \\
\cline{2-4}
(Ours) & SFH & 87 & 1194889.74 &  &  \\
\hline\hline
\multicolumn{3}{|l|}{Energy Consumption GT \cite{IMMMdata}}& 2056013.38 & 191779982 & / \\
\hline
\multicolumn{3}{|l|}{\multirow{2}*{Energy Estimation Accuracy Boosted by}} & \multicolumn{2}{|c|}{\textbf{Our Reconstruction Task}}& \color{Green}22.00 $\uparrow$ \\
\cline{4-6}
\multicolumn{3}{|l|}{} &\multicolumn{2}{|c|}{\textbf{Our Downstream Task}}& \color{Green}2.12 $\uparrow$ \\
\hline
\end{tabular}
\end{center}
\caption{\textbf{Experiment Results.} We tried our method with single archetype generation for MFH and SFH in the region Rancho Palos etc.}
\label{sheet:single}
\end{table*}

\begin{table*}
\begin{center}
\begin{tabular}{|l|c|c|c|c|c|c|}
\hline
\multirow{2}*{Region} & Energy Consumption & PBM \cite{doe_prototype} & \multicolumn{2}{|l|}{MARL with Only} & \multicolumn{2}{|l|}{MARL Restricted by}\\
~ &  GT\cite{IMMMdata}$(kWh)$ & Accuracy$(\%)$ & \multicolumn{2}{|l|}{Reconstruction Task $(\%)$} & \multicolumn{2}{|l|}{DTP $(\%)$} \\
\hline\hline
Rancho Palo etc. & 191779982 &  71.62 & 90.36 & \color{Green}18.74 $\uparrow$ & \textbf{91.08}  & 18.74 $\uparrow$ + \color{Green}0.72 $\uparrow$\\
\hline
Long Beach etc. & 104117941 &  73.10 & \textbf{98.96} & \color{Green}25.86 $\uparrow$ & 97.58  & 25.86 $\uparrow$ - \color{Bittersweet}1.37 $\downarrow$\\
\hline
Manhattan Beach etc. & 121545524 &  70.51 & 90.43 & \color{Green}19.92 $\uparrow$ & \textbf{92.88}  & 19.92 $\uparrow$ + \color{Green}2.45 $\uparrow$\\
\hline\hline
SUM & 417443447 &  71.66 & 92.52 & \color{Green}20.86 $\uparrow$ & \textbf{93.23} & 20.86 $\uparrow$ + \color{Green}0.70 $\uparrow$\\
\hline
\end{tabular}
\end{center}
\caption{\textbf{Experiment Results on Multiple Regions.} Our estimation compared with PBM using multiple archetypes.}
\label{sheet:multiple}
\end{table*}

\subsection{Ablation experiments}
In this experimental section, we would like to answer the following questions: (1) To what extent can our MARL framework improve the accuracy of current UBEM in a fully self-supervised manner with only reconstruction tasks? (2) Whether our MARL framework, when constrained by downstream tasks with a more disciplinary scope, can further improve our accuracy of UBEM?

To answer these two questions, we evaluated the performance of our method with or without DTP in each of the three neighborhoods of Los Angeles County as shown in Figure \ref{fig:region_map}: (b) Manhattan Beach, etc., (c) Rancho Palo, etc., and (e) Long Beach, etc. We input the archetype derived from our method into the energy simulations and compare the results to the one derived from the PBM \cite{doe_prototype}.

As residential structures represent a significant portion of energy consumption in densely populated urban zones, our analysis concentrates on these residential areas, particularly single-family housing (SFH) and multi-family housing (MFH). We feed the data from residential buildings across various regions into our trained encoder, thereby securing the latent representation for each structure.

For each representative footprint within the clustered building stock, we construct a building archetype for energy estimation. We use the building height and footprint to reconstruct the building envelope. Other specifications, such as the window-wall ratio and material are in alignment with the PBM building archetype from zone 3C \cite{doe_prototype}. Following the archetype construction, we perform energy simulation under Los Angeles International Airport climate data as extracted from \cite{epwweb}.

\begin{figure}[t]
\begin{center}
   \includegraphics[width=1.0\linewidth]{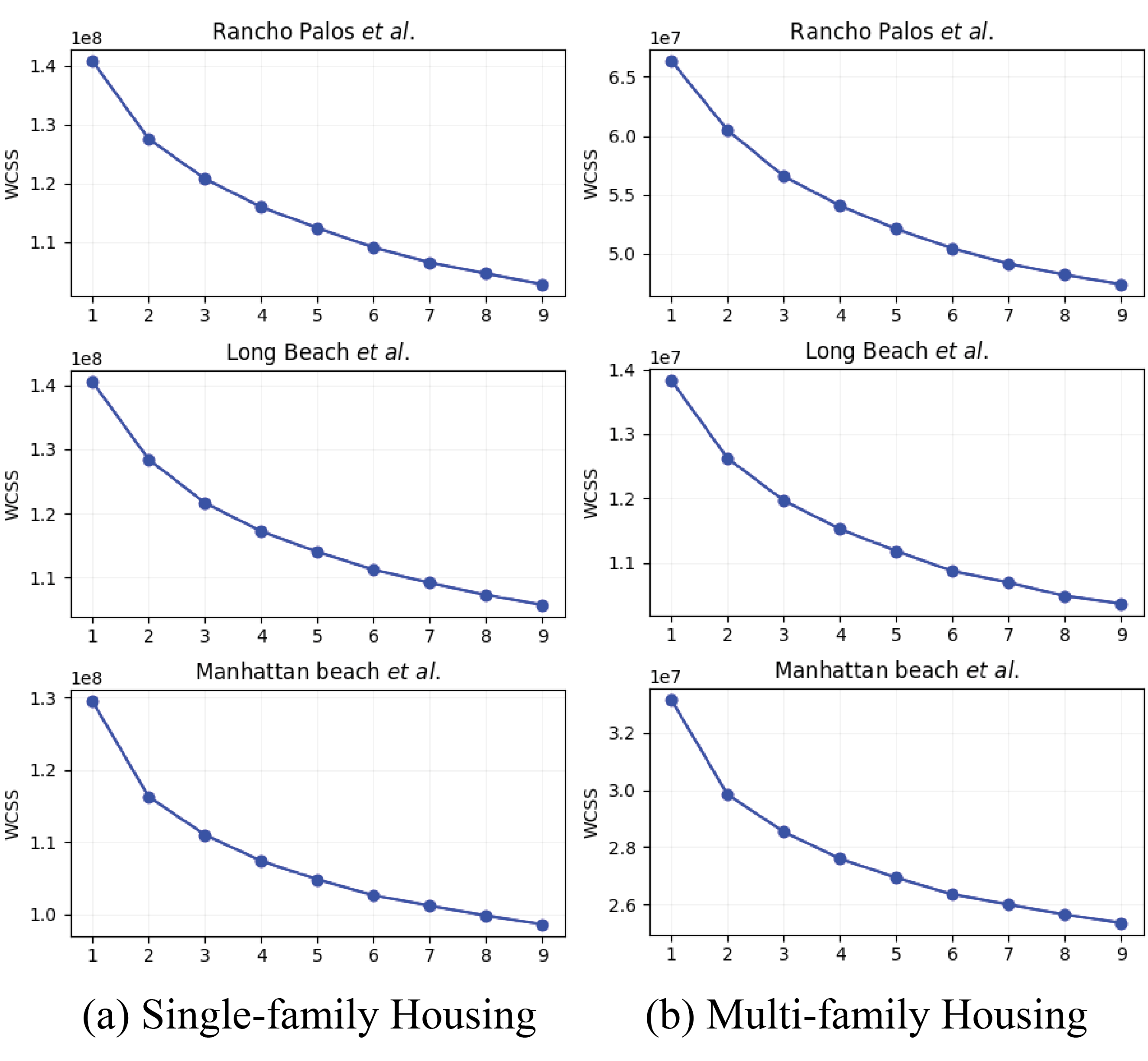}
\end{center}
   \caption {Within-cluster sum of squares for k-means clustering on latent space for different neighborhoods.}
\label{fig:kmeanswcss}
\end{figure}

\begin{figure}[t]
\begin{center}
   \includegraphics[width=1.0\linewidth]{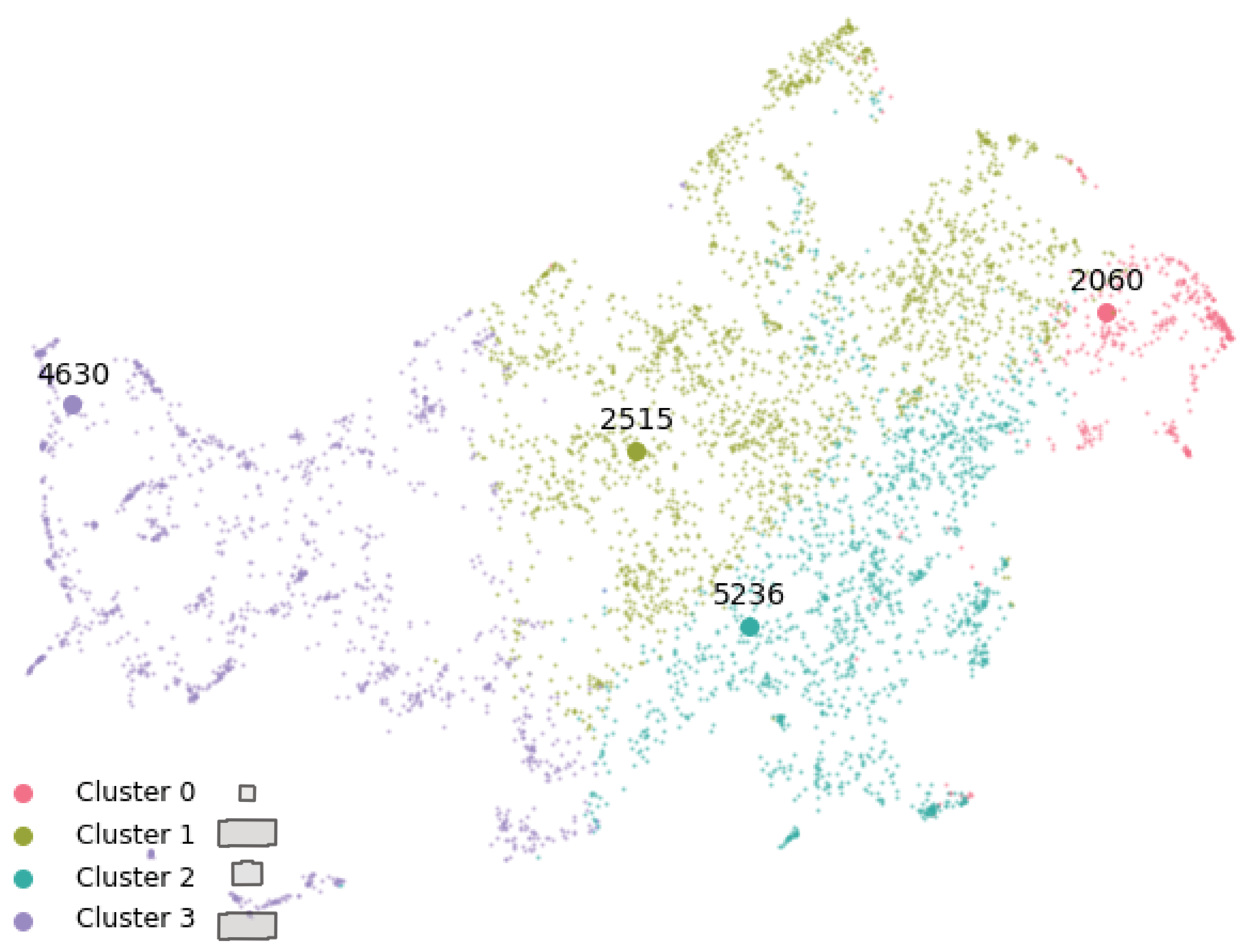}
\end{center}
   \caption{UMAP visualization of k-means clustering on latent space of region (c) with corresponding cluster center}
\label{fig:umap}
\end{figure}

\begin{table*}[t]
\begin{center}
\begin{tabular}{|l|c|c|c|c|c|c|}
\hline
\multirow{2}*{Region} & GT\cite{IMMMdata} & \multicolumn{2}{|c|}{PBM \cite{doe_prototype}} & \multicolumn{3}{|c|}{MARL+DTP (Ours)}\\
\cline{3-7}
~ & $(kWh)$ & $(kWh)$ & $(\%)$ & $(kWh)$ & \multicolumn{2}{|c|}{$(\%)$} \\
\hline\hline
Downey etc. & 187349182 &  144685496 & 77.23 & 201129362 & \textbf{92.64}  & \color{Green}15.42 $\uparrow$\\
\hline
Santa Monica etc. & 211891201 &  151819183 & 71.65 & 192191917 & \textbf{90.70}  & \color{Green}19.05 $\uparrow$\\
\hline
\hline
SUM & 399240383 &  296504678 & 74.27 & 393321279 & \textbf{98.52}  & \color{Green}24.25 $\uparrow$\\
\hline
\end{tabular}
\end{center}
\caption{\textbf{Open Set Experiment Results.} Energy consumption estimation in unseen regions. }
\label{sheet:unseen}
\end{table*}

\subsubsection{Single archetype}
The archetypes provided by PBM only contain one each for SFH and MFH \cite{doe_prototype}. So we first compute one center of the latent vectors for SFH and one for MFH respectively in Rancho Palos. Taking SFH as an example, when we obtain the center of the latent vectors corresponding to all the buildings in Rancho Palos, we input the nearest samples adjacent to that center as archetypes into the UBEM to obtain the average energy consumption of that region. This average energy consumption is weighted and summed over the building area to get the energy consumption of the whole region.

Table \ref{sheet:single} shows the result of this experiment. When there is only the image reconstruction task, the UBEM accuracy from our MARL-provided archetypes is 22 percent higher than the one from PBM's, and when the downstream tasks are added during model training, our method is 24.12 percent more accurate than the traditional method.

In the training of our model, our method is not supervised by the ground truth of energy consumption, and the whole process is only supervised by the self-supervision from the building footprints and the labeled supervision from the building meta information. However, it can be seen that when we consider the geometric features and architectural attributes of local buildings in a certain region, our method is able to learn more valuable representations and produce more locally characterized archetypes, which leads to a more accurate simulation of building energy consumption.

\subsubsection{Multiple archetype}

Based on our trained model, we use k-means clustering to the dimensionally reduced latent space to find representative building footprints prevalent in the selected neighborhoods. To evaluate the clustering results and determine the optimal number of clusters, the within-cluster sum of squares (WCSS) is used as an inertia measure. We select the number of clusters based on the elbow of WCSS, implying the optimal cluster number is within 2 to 5 (Figure \ref{fig:kmeanswcss}). 
With SFH numbers reaching 5537, 5317, and 5510, significantly surpassing the MFH counts of 1422, 2419, and 554 in Regions (b), (c), and (e), respectively, we opt for 4 archetypes among SFH and 2 among MFH for this experiment. 


We designate the center of each cluster as the representative building for that particular group. An example of the clustered latent space, along with the representative footprint, is visualized by Uniform Manifold Approximation and Projection (UMAP) in Figure \ref{fig:umap}.

We further tried our algorithm on three regions, Table \ref{sheet:multiple} shows our results. The archetypes derived from our algorithm still perform significantly better than the conventional archetype in terms of accuracy on UBEM. And the downstream tasks' restriction further improves the performance of our model in general.

\subsection{Open set}
Energy modeling for all buildings in a region is costly, not all regions have energy consumption data or simulation for each building, and most regions can only use one archetype for each building category \cite{doe_prototype}, which is designed at a country-wide scale. From this perspective, our approach is valuable and efficient because it can provide locale-specific building types, and does not require all building energy data as labels to oversee the entire model training process. We can train on any piece of area as long as GIS data is available.

We can even directly use our trained model for encoding building footprints in an open set. The experimental results presented in Table \ref{sheet:unseen} show the performance of our model in two unseen LA regions: (a) Santa Monica etc. and (d) Downey etc.  Our experimental results demonstrate that despite not being trained on these regions at all, our model still performs remarkably well in encoding new architectural footprints, leading to excellent representations and archetypes that enable it to succeed in the UBEM tasks.


\section{Conclusion}
In this research, we introduce Multi-scale Archetype Representation Learning (MARL) for the automated creation of building archetypes adaptable to diverse regions. This methodology incorporates building geometry into the building archetype construction process. Utilizing representation learning and downstream task restriction, we extract implicit features from building stock and then employ k-means clustering to identify representative building footprints as archetypes. We further refine our approach by constraining the footprint reconstruction process with energy performance-related metadata such as vintage year and building use type. 

As a result, we propose an automated building archetype construction method. We validate the efficacy of our model by benchmarking its energy estimation performance against conventional building archetypes, both for seen and unseen neighborhoods. Our experimental outcomes demonstrate that our approach outperforms conventional models, accentuating the potential of our method in enhancing the precision and adaptability of Urban Building Energy Modeling, and presenting an advancement in the domain of building archetype development and building energy modeling.

By integrating building geometries into the archetype construction process, architects and urban planners can make informed choices about neighborhood configurations, building orientations, and other design elements impacting energy consumption. Our method can also identify areas with suboptimal energy performance, indicating opportunities for retrofitting and design modifications. Furthermore, by focusing on locale-specific geometric nuances, we offer a nuanced approach to building archetypes, fostering designs that harmonize with local characteristics. This promotes energy efficiency and locale-specific design, and provides insights into the wider implications of building morphology on energy and environmental factors.


{\small
\bibliographystyle{ieee_fullname}

}

\end{document}